\documentclass[conference]{IEEEtran}
\IEEEoverridecommandlockouts
\usepackage{cite}
\usepackage{amsmath,amssymb,amsfonts}
\usepackage{algorithmic}
\usepackage{textcomp}
\usepackage{xcolor}
\usepackage{graphicx}
\usepackage[caption=false,font=footnotesize]{subfig}
\usepackage{float}
\usepackage{makecell}
\usepackage{subcaption}
\usepackage{multirow}
\usepackage{hyperref}

\def\BibTeX{{\rm B\kern-.05em{\sc i\kern-.025em b}\kern-.08em
    T\kern-.1667em\lower.7ex\hbox{E}\kern-.125emX}}

\DeclareRobustCommand{\IEEEauthorrefmark}[1]{\smash{\textsuperscript{\footnotesize #1}}}

\begin{document}

\title{Simulating Safe Bite Transfer in Robot-Assisted Feeding with a Soft Head and Articulated Jaw
\thanks{
    The research was conducted at the Future Health Technologies at the Singapore-ETH Centre, which was established collaboratively between ETH Zurich and the National Research Foundation Singapore. This research is supported by the National Research Foundation Singapore (NRF) under its Campus for Research Excellence and Technological Enterprise (CREATE) programme.
}

\thanks{
\IEEEauthorrefmark{1}School of Mechanical and Aerospace Engineering, Nanyang Technological University Singapore
}
\thanks{
\IEEEauthorrefmark{2}Singapore-ETH Centre, Future Health Technologies Programme
}
\thanks{
    Corresponding authors' e-mails: \{sherwins001, janne001\}@e.ntu.edu.sg
}
}

\author{
    \IEEEauthorblockN{
        Yi Heng San\IEEEauthorrefmark{1,*},
        Vasanthamaran Ravichandram\IEEEauthorrefmark{1,*},
        J-Anne Yow\IEEEauthorrefmark{1,2,*}, 
        Sherwin Stephen Chan\IEEEauthorrefmark{1,*},
    }
    \IEEEauthorblockN{
        Yifan Wang\IEEEauthorrefmark{1},
        Wei Tech Ang\IEEEauthorrefmark{1,2}
    }
    \IEEEauthorblockA{
        \IEEEauthorrefmark{*}These authors contributed equally
    }
}

\maketitle

\begin{abstract}
Ensuring safe and comfortable bite transfer during robot-assisted feeding is challenging due to the close physical human-robot interaction required. This paper presents a novel approach to modeling physical human-robot interaction in a physics-based simulator (MuJoCo) using soft-body dynamics. We integrate a flexible head model with a rigid skeleton while accounting for internal dynamics, enabling the flexible model to be actuated by the skeleton. Incorporating realistic soft-skin contact dynamics in simulation allows for systematically evaluating bite transfer parameters, such as insertion depth and entry angle, and their impact on user safety and comfort. Our findings suggest that a straight-in-straight-out strategy minimizes forces and enhances user comfort in robot-assisted feeding, assuming a static head. This simulation-based approach offers a safer and more controlled alternative to real-world experimentation. Supplementary videos can be found at: \url{https://tinyurl.com/224yh2kx}.

\end{abstract}


\section{Introduction}
Robot-assisted feeding systems offer a promising solution for improving the quality of life of individuals requiring feeding assistance while also alleviating caregiver burden. However, implementing safe and effective bite transfer -- the process of delivering food from a utensil into a user's mouth -- presents unique challenges due to the close, physically interactive nature of the task. 

Autonomous bite transfer has been proposed, with earlier works detecting the user's mouth pose and positioning the food in front of a care recipient's mouth \cite{park2016towards, gallenberger2019transfer}. However, leaning forward for an outside-mouth bite transfer can be impossible or tiring for those with more severe upper body and neck mobility limitations, prompting recent works to study in-mouth bite transfer. 
To further enhance user comfort and safety, compliant controllers based on force feedback have been implemented, adjusting compliance during entry and exit phases \cite{shaikewitz2023mouth} or classifying physical interactions using visual and haptic feedback to select a suitable controller \cite{jenamani2024feel}.
Despite these advancements, current methods generally assume a fixed, pre-planned path and rely solely on force feedback to adapt to contact forces without adjusting bite transfer parameters such as insertion depth, exit angle, or exit depth.

Ensuring user safety is paramount in developing robot-assisted feeding systems, as direct physical contact risks excessive force, resulting in discomfort or even injury to sensitive areas like the mouth. Exploring different ways to perform bite transfer with real users is challenging due to ethical and safety concerns. 


\begin{figure}[t]
    \centering
    \includegraphics[width=\linewidth]{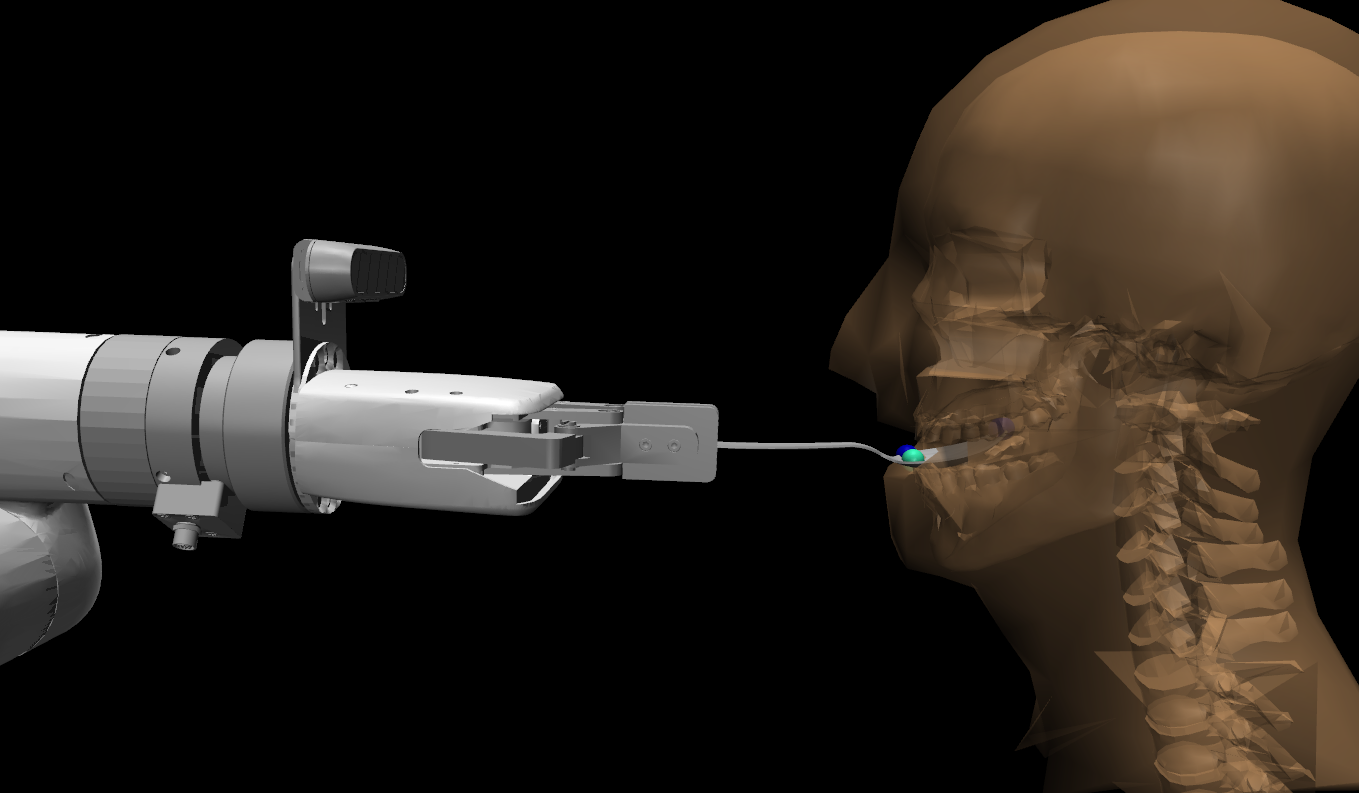}
    \caption{Physical human-robot interaction simulation of bite transfer with contact forces between the spoon and the soft-body mouth shown by the blue and green spheres. The opacity of the head and skull is decreased to show the contact points. 
    }
    \label{fig: bite-transfer-sim}
\end{figure}

Human-in-the-Loop (HITL) simulation has been applied to assistive tasks that demand a high level of safety, such as assisted dressing \cite{kapusta2019personalized} and exoskeleton control \cite{kumar2020learning}. These simulations have proven valuable in enabling the development and evaluation of robotic controllers that involve physical Human-Robot Interaction (pHRI) and in some cases, facilitated the sim-to-real transfer of robotic controllers, particularly for complex pHRI tasks, such as learning and optimizing exoskeleton control \cite{luo2024experiment}. However, most existing approaches rely on rigid-body skeletal \cite{chan2023creation}, musculoskeletal \cite{luo2024experiment}, or primitive-shaped human models \cite{clegg2020learning}, which lack the ability to accurately model detailed physical interactions, particularly at the contact level. This limitation poses a significant challenge for tasks requiring precise contact modelling, such as robot-assisted feeding.

\begin{figure*}[t]
    \centering
    \includegraphics[width=0.98\linewidth]{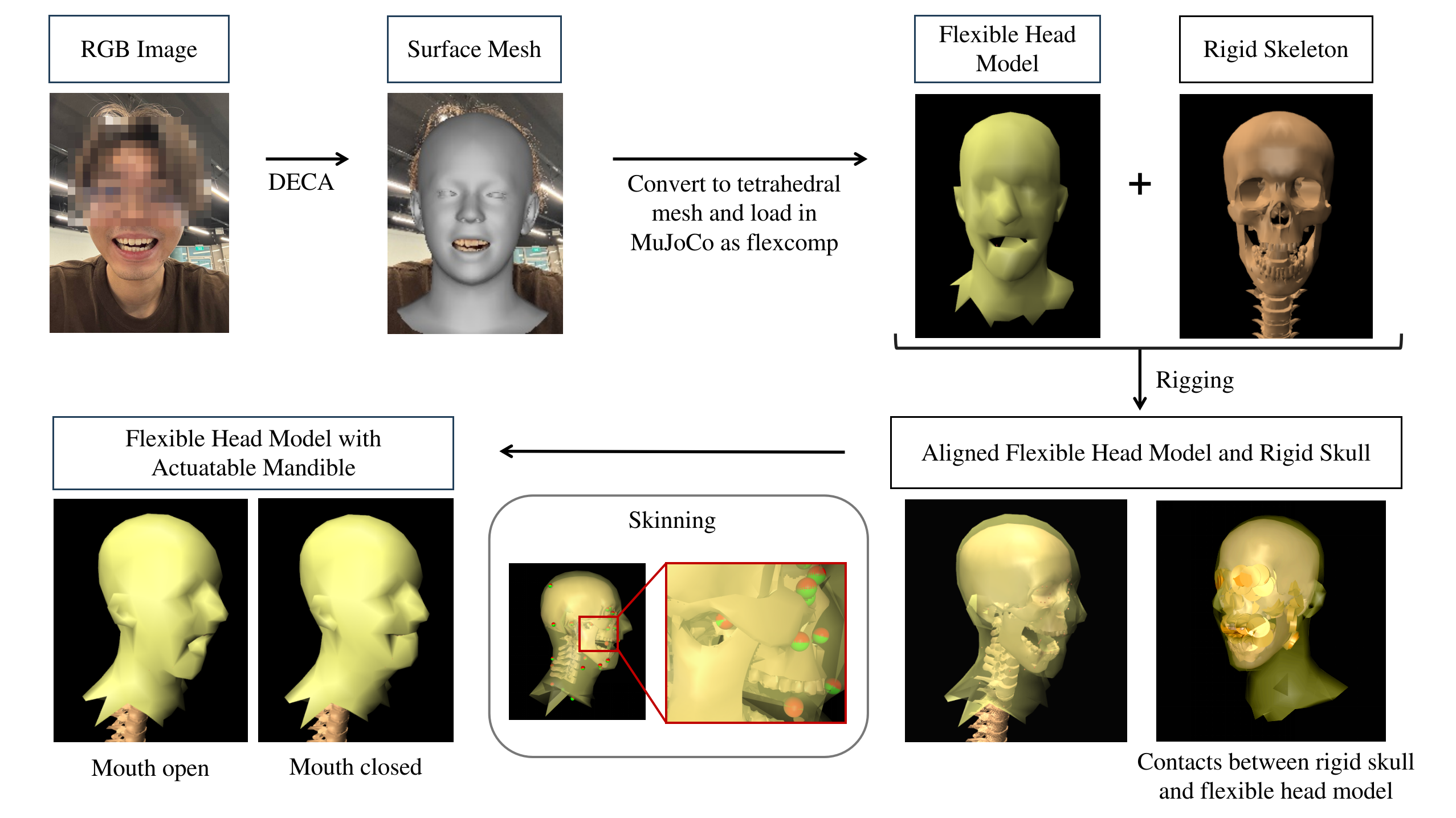}
    \caption{Overview of our approach to model a head with soft-body dynamics. (Top) A flexible head model is generated from an RGB image through surface mesh reconstruction, tetrahedral meshing, and loading into MuJoCo as a flexible element. (Bottom) The flexible head model is combined with a rigid skeleton through rigging, and skinning is applied to bind the two. Contacts are detected and marked with yellow cylinders (bottom right), after which sites are created at the detected points, visualized using red and green balls (bottom center). Tendon elements then connect the sites to form a fully rigged and actuated head model. This setup enables realistic deformation, mandible actuation, and mouth movement simulations.}
    \label{fig: overview}
\end{figure*}

In robot-assisted feeding, particularly the bite transfer process, research in pHRI simulation remains sparse \cite{erickson2020assistive, ye2022rcare}. This gap is largely attributed to the absence of a high-quality face and mouth model capable of accurately representing the shape and motions involved in chewing and biting. To date, the only physics-based simulator attempting to integrate soft-body human modelling for assistive robotics is RCareWorld \cite{ye2022rcare}. However, its human models are constrained to predefined configurations with fixed shape and mobility parameters, limiting personalization to individual users. While skinned biomechanical human models, such as those described in \cite{keller2023skin}, have introduced deformable skin, they have largely overlooked the internal dynamics between the skin and the underlying skeleton in a physics-based simulation. Similarly, our previous approach \cite{loke2023personalised}, which manually rigged the skin and skeleton as a unified whole-body model using rigid \texttt{pin} elements, further highlighted these limitations. Although functional, this method introduced redundant contact forces during interactions, underscoring the need for a more sophisticated approach to model soft-tissue dynamics and natural interactions. These limitations highlight the need for a high-fidelity simulation framework capable of accurately modelling soft-body dynamics and human features, particularly the face and mouth, to enable detailed investigation of contact forces and motion dynamics in tasks such as robot-assisted feeding.


To address these gaps, we present a novel framework for simulating bite transfer using a soft-body human head model with an articulated mandible developed in MuJoCo \cite{todorov2012mujoco}. This model enables the analysis of how variations in bite transfer parameters impact force distribution on the user, facilitating systematic experimentation without safety risks. By simulating these interactions, we aim to identify optimal bite transfer strategies that maximize user comfort by minimizing contact forces. Key contributions of this work include: 
\begin{enumerate}
    \item A method for creating a realistic, flexible human head model with a movable mandible from an RGB image, enabling the simulation of mouth opening and closing by controlling the joint of the rigid mandible.
    \item Novel evaluation and analysis of bite transfer parameters to minimize contact forces and enhance user comfort.
\end{enumerate}
These contributions provide a foundation for improving safety in robot-assisted feeding systems and modelling physical human-robot interactions more accurately. 

\section{Approach}
To facilitate bite transfer parameter analysis in simulation, we developed a human head model in MuJoCo featuring soft-body contact dynamics. Using the \texttt{flexcomp} body element in MuJoCo, we constructed a volumetric, deformable head model that mimics the mechanical properties of human skin and tissue. The head model was rigged with a rigid skull structure, enabling mandible actuation to simulate realistic mouth opening and closing motions. This section provides a detailed explanation of the human head model creation process.

\subsection{Generation of Flexible Head Model}
\label{sec: soft-body}
We take an RGB image of a human head and process it using DECA \cite{DECA:Siggraph2021} to reconstruct a facial model, generating a rigid surface mesh of the head (Fig. \ref{fig: overview}). This surface mesh is then converted into a high-quality tetrahedral mesh using the method described in \cite{hu2020fast}, which incrementally inserts triangles to create a volumetric mesh. The conversion to a tetrahedral mesh is necessary for compatibility with MuJoCo, which requires a volumetric representation of the mesh to create the necessary relationships across all vertices for soft-body dynamics. To improve computational efficiency, the mesh is simplified to produce a final model with 342 vertices.

To define the soft body properties in simulation, the \texttt{flexcomp} elasticity attributes in MuJoCo were utilized to represent the mechanical behaviour of deformable materials accurately. The parameters used -- Young's modulus and Poisson's ratio -- were selected based on a combination of literature review of human soft tissue material \cite{Singh_2021, ARNOLD2023286} (Table \ref{tab: human-tissue}) and the empirical testing of the functional requirements of the model. We used a Young's Modulus of $5\times10^4$ and a Poisson Ratio of $0.4$. To ensure realistic mass distribution, we assigned a total mass of 5 kg to the flexible head model, consistent with the plausible range of human head mass reported in the literature \cite{yoganandan2009physical}. This mass was evenly distributed across all vertices of the flexible head, while the mass of the rigid skull was set to near zero to avoid redundancy and ensure that the flexible head accurately represents the primary source of soft tissue mass.

\begin{table}[h]
    \centering
    \caption{Material Properties of Human Soft Tissues}
    \begin{tabular}{|l|l|l|}
        \hline
        Tissue Type & Poisson Ratio & Young's Modulus/kPa \\ \hline
        Connective Tissue & 0.3 & $2.25\times10^5$ - $1.5\times10^6$ \\ \hline
        Muscle & 0.3 - 0.493 & 39 - 164 \\ \hline
        Fat & 0.13 - 0.5 & 18 - 24 \\ \hline
    \end{tabular}
    \label{tab: human-tissue}
\end{table}

\subsubsection{Rigging the Head Model}
To enable mouth opening and closing movements, we utilized a rigid-body head and neck model from OpenSim \cite{mortensen2018inclusion} and converted it to a MuJoCo model \cite{ikkala2022converting}. The rigid skull consists of two components -- the upper skull and the mandible. We created a six-degree-of-freedom (DoF) joint between these two bodies to simulate realistic mandible movement. We manually align the position of the skull inside the flexible head model due to the differences in the initial orientation.

\subsubsection{Skinning}
Skinning refers to the process of binding the flexible head model to the rigid skull by associating regions of the head's surface with specific parts of the skull, ensuring that movements of the skull result in corresponding deformations of the skin, as seen in the mouth opening and closing in Fig. \ref{fig: overview}. To model the internal dynamics between the soft skin and the skull, we mimicked the natural anatomical attachment of soft tissues to the underlying skeletal structure. Ideally, the flexible head would be attached to underlying muscles; however, in the absence of detailed muscle models for the skull, the flexible head was directly attached to the skeleton. In this context, the \texttt{tendon} element in MuJoCo was employed to represent fascia and connective tissues, which exhibit viscoelastic properties. These properties facilitate the independent yet cohesive motion of the skin in response to movements of the mandible. The \texttt{tendon} was created according to assigned pairs of \texttt{site} attributes on both the skin and the skull. All interaction forces between the skin and the skull were transmitted exclusively through the defined \texttt{tendon} elements.

To determine the appropriate locations for the \texttt{tendon} elements, we analyzed the natural contact points between a rigid skull and a volumetric flexible head. After rigging the rigid skull and volumetric flexible head in MuJoCo, contact interactions between the surface of the skull and internal volume of the flexible head were identified. These contact points served as the primary indicators of potential locations where structural interactions between the skin and the skull should occur, ensuring stability during simulation runtime. Based on these contact points, the tendon elements were positioned by identifying the closest vertices of the soft head model to the contact points, effectively aligning the tendons with areas of natural contact.

With the detected set of contact points $C = \{c_1, c_2, \dots, c_n\} \subset \mathbb{R}^3$ and the vertex set of soft body head model $V = \{v_1, v_2, \dots, v_n\} \subset \mathbb{R}^3$, a linear search algorithm was applied to identify the subset of vertices $V_{closest} \subseteq V$ nearest to each contact point. To ensure the uniqueness of \texttt{tendon} element in each interaction area, the algorithm built an injective mapping $M: C \to V_{closest}
$ by iterating through each contact point and calculating the Euclidean distance between that point and each vertex, represented as following:
\begin{equation}
M(c_i) = \arg\min_{v_j \in V \setminus V_{\text{closest}}} \|c_i - v_j\|_2, \quad \text{where } c_i \in C
\end{equation}

For each $v_k \in V_{closest}$, a pair of \texttt{site} attributes were created at the coordinates of $v_k$ and assigned to both the rigid skull and flexible head. The \texttt{tendon} elements were used to link each pair of corresponding \texttt{site} attributes. These \texttt{tendon} elements were configured with adjustable viscoelastic properties and allowable stretching ranges, which were empirically determined through testing and refinement to ensure realistic behavior. Once the \texttt{tendon} elements were activated, default contact interactions between the rigid skull and flexible head were disabled. This ensured that the motion of the flexible head was driven exclusively by the \texttt{tendon} connections, with all interaction forces transmitted solely through these elements. By relying on tendon-driven dynamics, the head model exhibited predictable and controlled movements, entirely driven by the motion of the mandible.
The developed skinning method is novel in that it addresses the internal dynamics between the flexible head and skull, enabling realistic motion actuation and force transmission.
In our application, this integration enables precise control of the head model, facilitating detailed simulations of bite transfer in robot-assisted feeding systems. We believe that this method has the potential to be extended to other body parts in future work, enabling applications beyond head modeling, such as full-body modeling and biomechanical interaction studies, to support a broader range of research and simulation contexts.

\subsection{Modifying In-Mouth Bite Transfer Trajectories}
\begin{figure}
    \centering
    \subfloat[Entry angle, \(\alpha\) ]{%
        \includegraphics[width=0.48\columnwidth]{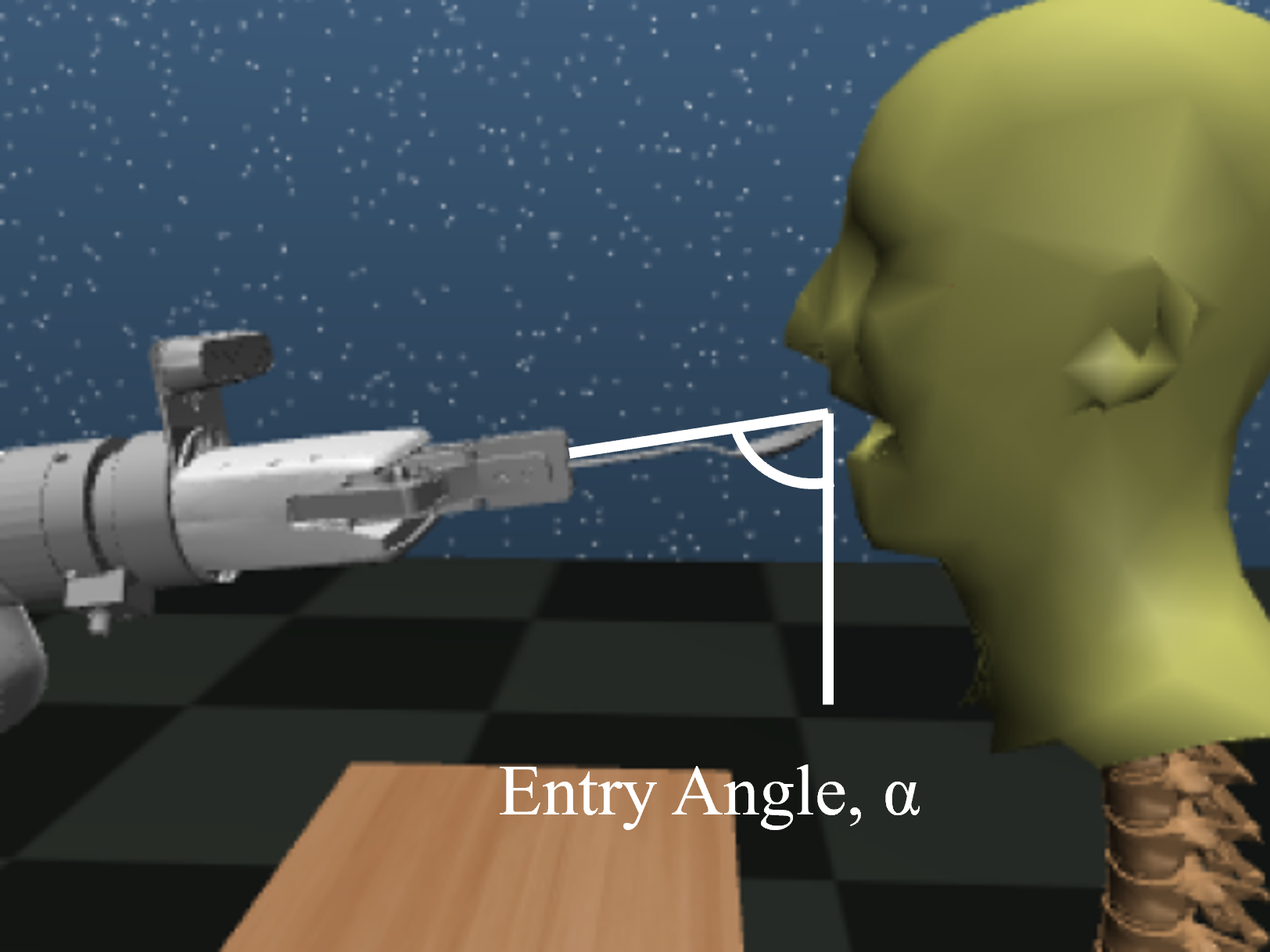}
        }
    \hfill
    \subfloat[Insertion depth,  \(d\) ]{%
        \includegraphics[width=0.48\columnwidth]{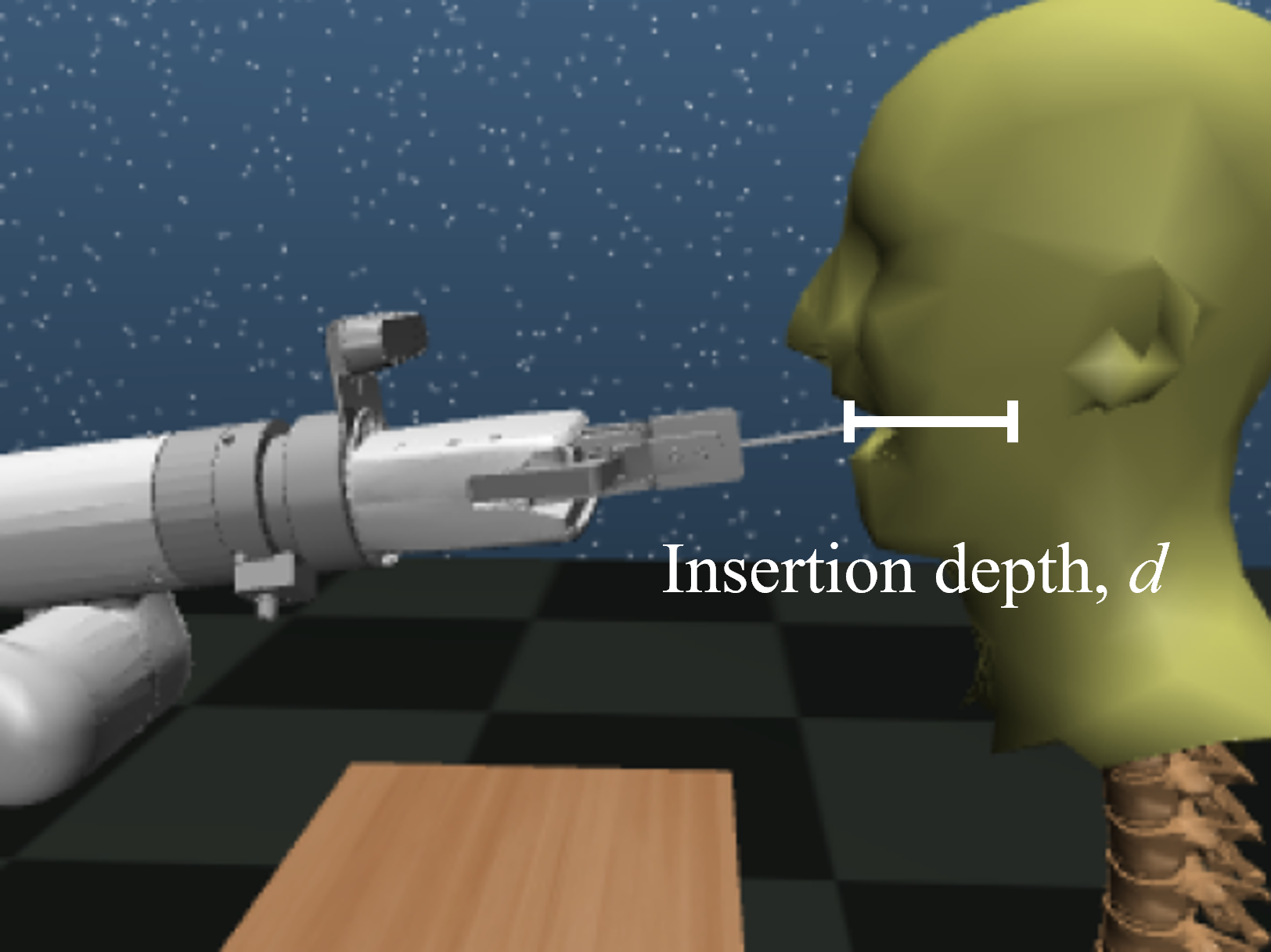}
        }
    \vspace{-1.0em}
    \subfloat[Exit angle, \(\beta\)]{%
        \includegraphics[width=0.48\columnwidth]{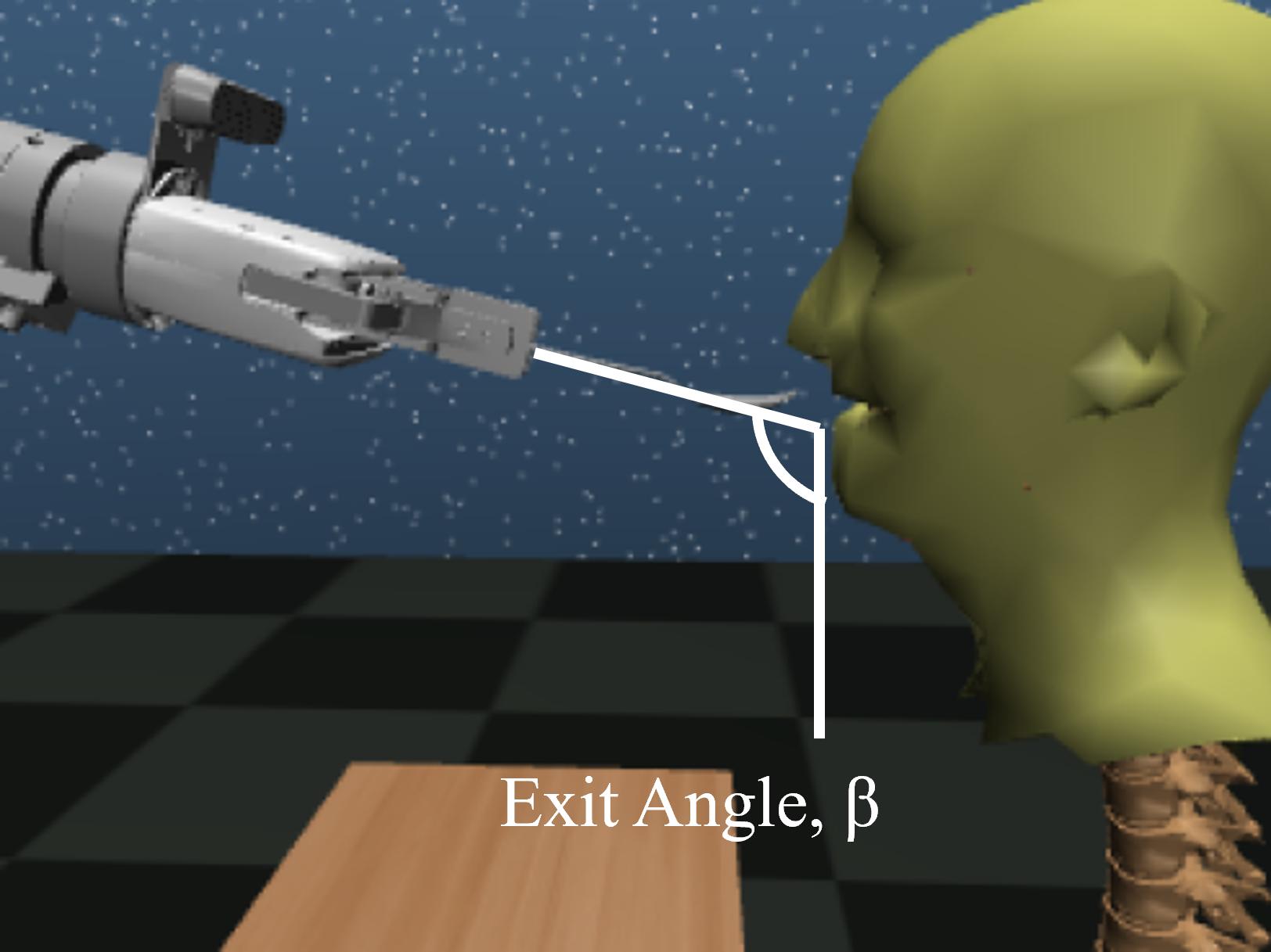}
        }
    \hfill
    \subfloat[Exit depth, \(e\)]{%
        \includegraphics[width=0.48\columnwidth]{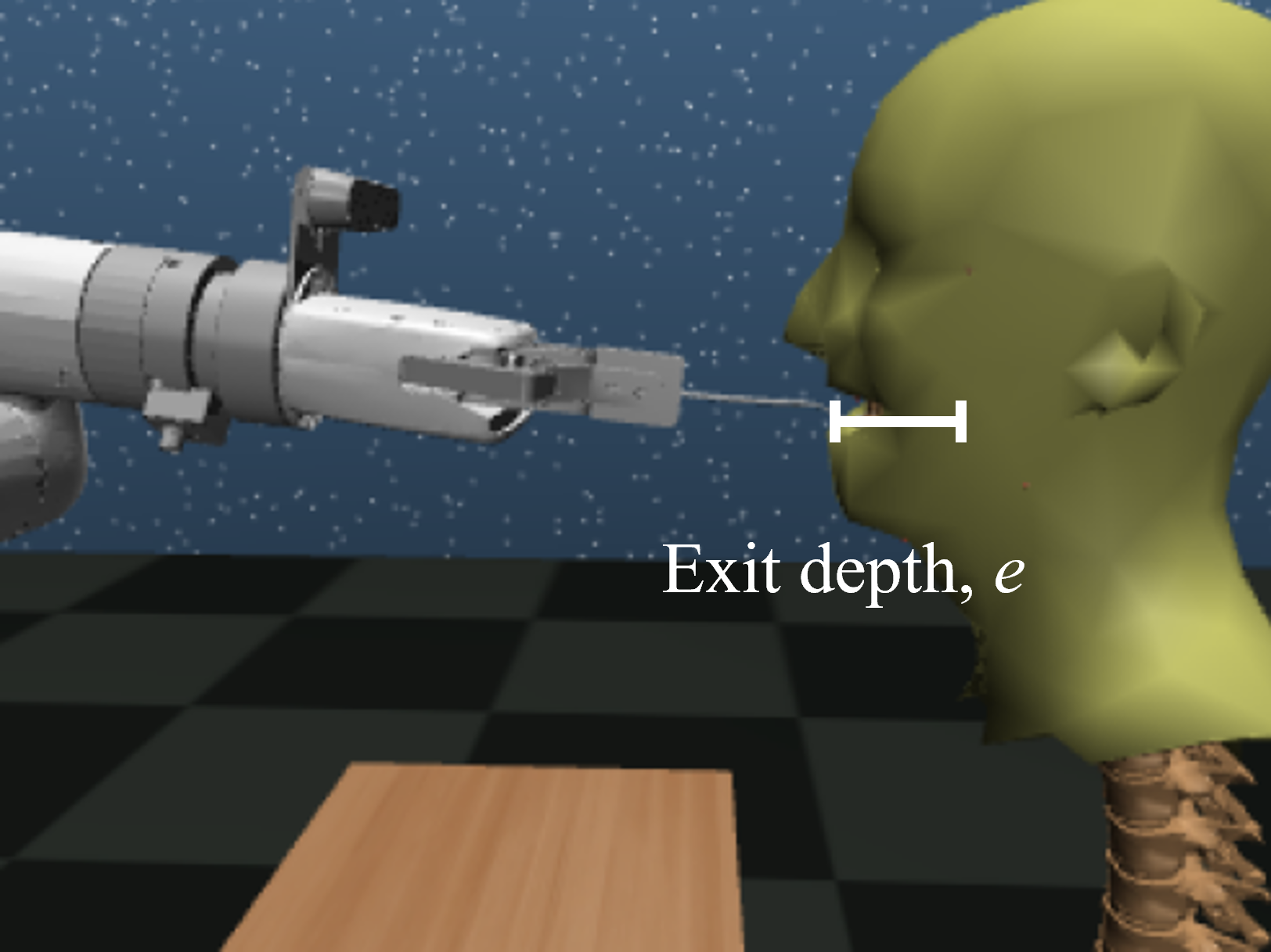}
        }
    \caption{Bite transfer parameters modified in our experiments}
    \label{fig: bite_transfer_parameters}
\end{figure}

To analyse the physical human-robot interaction forces exerted on the user by a spoon during bite transfer, we defined bite transfer as a parameterized trajectory, as shown in Fig. \ref{fig: bite_transfer_parameters}. 
In general, bite transfer can be divided into 2 phases, the entry phase and exit phase\cite{shaikewitz2023mouth}. 
We assume that bite transfer starts at a fixed initial pose \( p \), defined by a fixed distance of \(50 mm\) away from the mouth pose. As illustrated in Fig. \ref{fig: bite_transfer_parameters}, the spoon enters the user's mouth with an insertion depth \( d \) and entry angle \( \alpha \) during the entry phase. After the user closes the mouth and a bite is detected, the spoon exits the user's mouth with an exit depth \( e \) and exit angle \( \beta \). 

These parameters are then systematically adjusted within our simulation environment to observe their effects on contact forces exerted during bite transfer.

\section{Experiments and Results}
We conducted experiments in simulation to identify the optimal set of bite transfer parameters that maximise user comfort during bite transfer, defined here as minimising contact forces applied during the interaction. 

\subsection{Experiment Setup}   
We set up a simulated environment in MuJoCo \cite{todorov2012mujoco} with the xArm-6 robotic arm (UFactory) grasping a spoon and the head model (Fig \ref{fig: bite-transfer-sim}). We selected this robotic arm as we have access to the physical device for future experimental validation. However, our approach is not limited to this specific robot and can be adapted to other robotic arms. The starting pose of the robot arm is fixed throughout the experiment for consistency. After the spoon is inserted into the user's mouth in the entry phase, we actuated the mandible joint of the head model to simulate mouth closing. We assume the rest of the human body remains static, reflecting scenarios where patients have limited head mobility. This assumption simplifies the analysis and allows us to focus on the influence of bite transfer parameters on bite transfer interaction forces.

To evaluate the impact of different bite transfer parameters, we measured the contact forces applied during the bite transfer phase. As multiple contact points could exist per time step, we compute the average force exerted per time step, denoted as \( f_{t} \). The key metrics considered were:
\begin{itemize}
    \item \textbf{Peak contact force}: the maximum average force \( f_{t}^{max} \) recorded during the interaction
    \item \textbf{Total contact force}: the summation of the average forces \( \sum f_{t} \) applied throughout the bite transfer
\end{itemize}
The best parameters should minimize both peak and total contact forces. Peak force affects instantaneous user comfort, as high forces in a short duration can cause discomfort or injury. Total force, the cumulative force over time, impacts overall comfort as prolonged low-level forces that can still cause discomfort if sustained for extended periods.

We performed a series of trials to analyze bite transfer parameters, as detailed in Fig. \ref{fig: bite_transfer_parameters}. The parameters were first varied during the entry phase to identify the optimal settings, followed by similar trials for the exit phase. Table \ref{tab: bite_transfer_parameters} summarises the specific parameter ranges and increments.
These ranges were set based on human eating behaviour and the geometry of the spoon, where the spoon bowl length was \(60\)mm.

\begin{table}[t]
    \centering
    \caption{Bite Transfer Parameter Ranges and Increments}
    \label{tab: bite_transfer_parameters}
    \begin{tabular}{|c|c|c|c|}
        \hline
        \textbf{Phase}       & \textbf{Parameter}          & \textbf{Range}           & \textbf{Increment} \\ 
        \hline
        \multirow{2}{*}{Entry} & Entry Angle (\(\alpha\))    & 80° to 110°              & 10°                \\ 
                             & Insertion Depth (\(d\))     & 50 mm to 100 mm          & 10 mm              \\ 
        \hline
        \multirow{2}{*}{Exit} & Exit Angle (\(\beta\))      & 80° to 120°              & 10°                \\ 
                             & Exit Depth (\(e\))          & 0 mm to 40 mm       & 10 mm            \\ 
        \hline
    \end{tabular}
\end{table}

\subsection{Results and Analysis}
\begin{figure}[ht]
    \centering
    \includegraphics[width=0.85\linewidth]{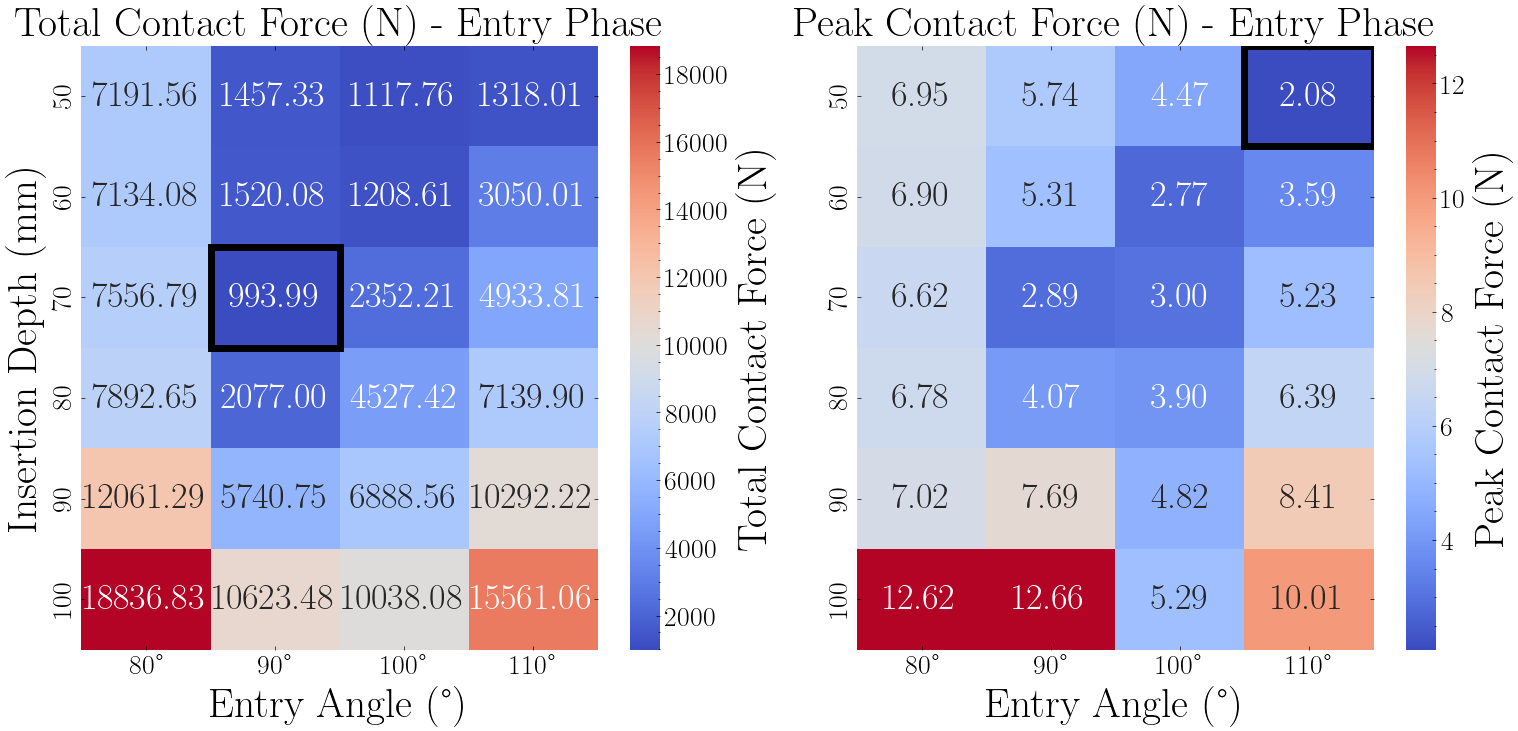}
    \caption{Force Analysis Results -- Entry Phase}
    \label{fig:total_contact_force_entry}
\end{figure}

\begin{figure}[ht]
    \centering
    \includegraphics[width=0.99\linewidth]{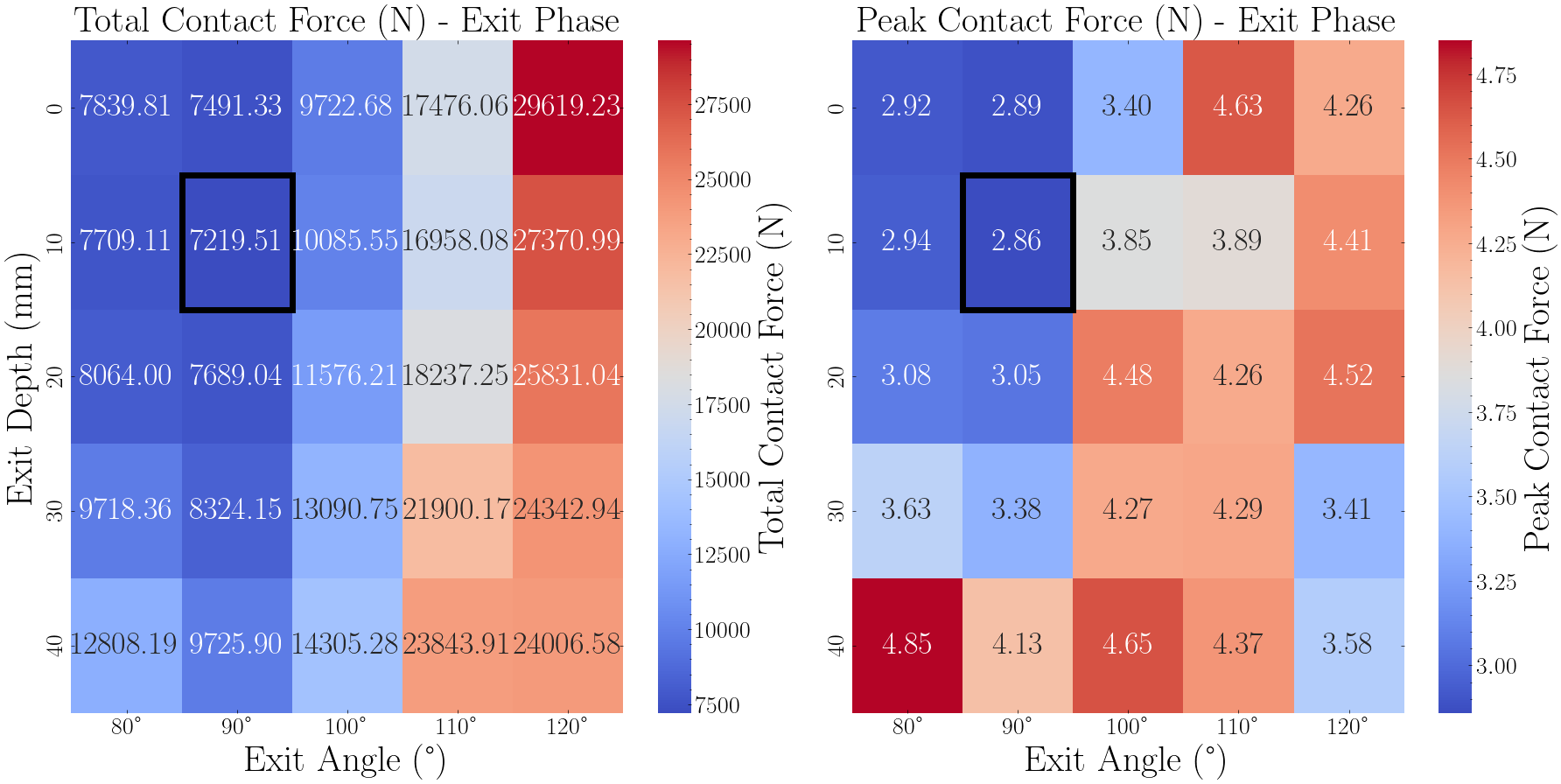}
    \caption{Force Analysis Results -- Exit Phase}
    \label{fig:total_contact_force_exit}
\end{figure}

The results of our experiments are shown in Figures \ref{fig:total_contact_force_entry} and \ref{fig:total_contact_force_exit}.

\begin{figure*}[t]
    \centering
    \begin{tabular}{cc}
        \includegraphics[width=0.48\textwidth]{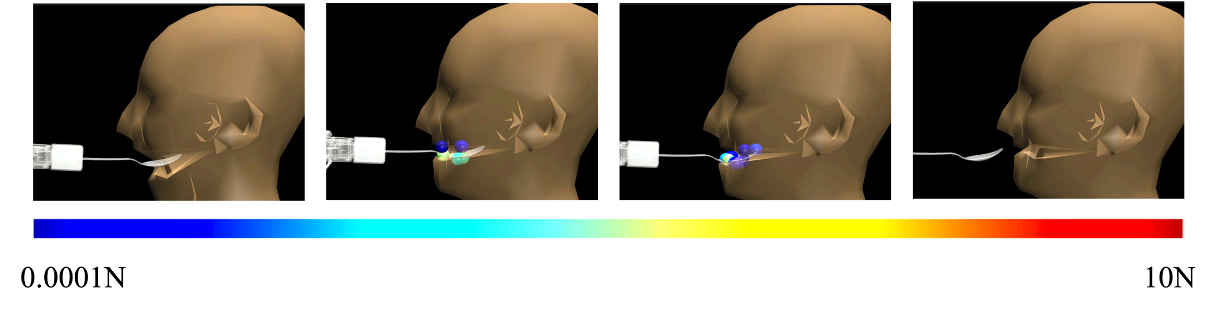} &
        \includegraphics[width=0.48\textwidth]{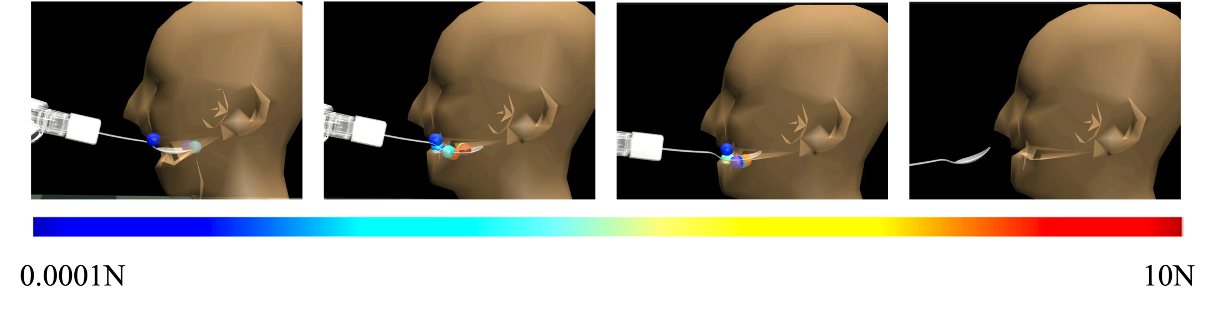} \\
        \includegraphics[width=0.46\textwidth]{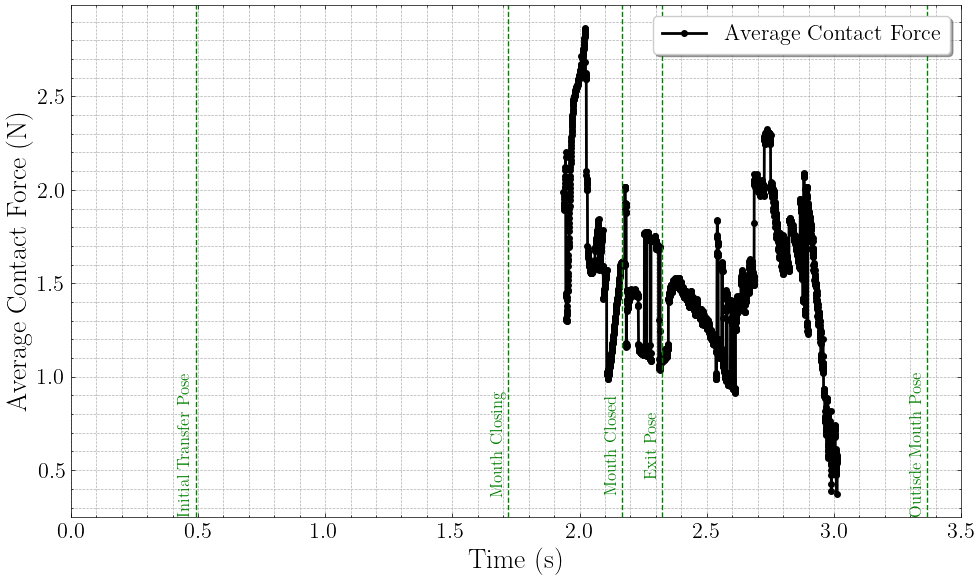} &
        \includegraphics[width=0.46\textwidth]{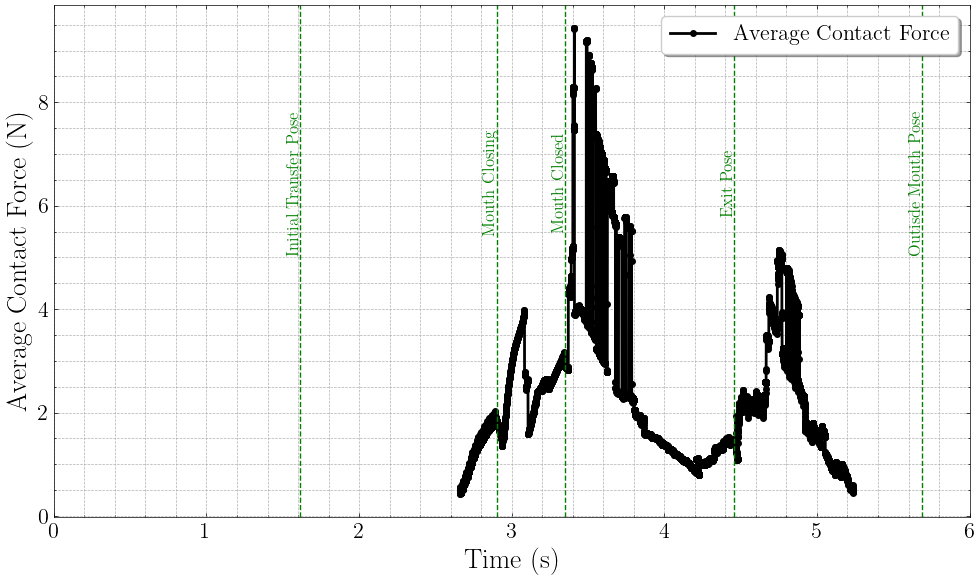} \\
        (a) Bite transfer with optimal parameters & (b) Bite transfer with suboptimal parameters \\
        (\(\alpha = 90\)°, \(d = 70 \)mm, \(\beta=90\)°, \(e = 10\) mm) & (\(\alpha = 100\)°, \(d = 80 \)mm, \(\beta = 80\)°, \(e = 40 \)mm)\\
    \end{tabular}
    \caption{Bite transfer in simulation: The top row shows screen captures of the bite transfer process, with contact forces visualized as coloured spheres, showing increased contacts in (b). The bottom row shows the corresponding contact forces over time. Note that the y-axis scales are different for clarity.}
    \label{fig: results}
\end{figure*}

\subsubsection{Entry Phase}
The simulation results show that both total and peak contact forces generally increase with greater insertion depths at entry angles of 100° and 110°. This is due to the larger contact surface between the spoon and the inside of the mouth as depth increases. 
The entry angle of 80° exhibits higher total and peak contact forces, especially at deeper depths, due to the tip of the spoon contacting the mouth more.

The lowest total contact force of 993.99N occurs at \(\alpha=90\)° and  \(d=70\)mm. Although the lowest peak contact force of 2.08N occurs at \(\alpha = 110\)° and \(d = 50\)mm, the peak force of \(\alpha=90\)° and  \(d=70\)mm is only slightly higher at 2.89N. This marginal increase in peak contact force of less than 1N is unlikely to be perceptible to a human. This minimal difference in peak contact force, combined with the substantially lower total contact force, makes \(\alpha=90\)° and  \(d=70\)mm the more optimal parameter set.
The optimal entry angle suggests the spoon should be perpendicular to the mouth, while the optimal insertion depth should be slightly longer than the spoon bowl to minimize contact forces.

As shown in Fig. \ref{fig: results}, the peak contact force for the entry phase occurs when the mouth is closing, which aligns with the expected increase in forces when the mouth interacts with the spoon. 



\subsubsection{Exit Phase}
For the exit phase, the results indicate a general trend where the peak contact force increases as the exit depth \(e\) increases. Additionally, exit angles closer to \(90\)° show lower total and peak contact forces. As the exit angle deviates further from \(90\)°, the transition from the exit depth pose to the exit angle pose results in more rotation within the mouth, leading to increased forces.




Based on the results, the parameter combination that minimizes the total contact force with the mouth and maximises the comfort of the user is an entry angle \(\alpha = 90\)°, an insertion depth \(d = 70\)mm, an exit angle \(\beta= 90\)° and an exit depth \(e=10\)mm. 
Screen captures of the bite transfer process for both the optimal set of parameters (\(\alpha = 90\)°, \(d = 70 \)mm, \(\beta=90\)°, \(e = 10\) mm) and a random set of suboptimal parameters (\(\alpha = 100\)°, \(d = 80 \)mm, \(\beta = 80\)°, \(e = 40 \)mm) are shown in Fig. \ref{fig: results} to better visualise the difference in contact forces for each stage of the bite transfer process.
For the optimal set of bite transfer parameters, the peak contact force occurs during the entry phase when the mouth is closing, while the peak contact force occurs during the exit phase for suboptimal parameters.

Thus, from our experiments, we observe that having a simple bite transfer process of ``straight-in, straight-out" given the same entry \(\alpha\) and exit angle \(\beta\) of 90° minimises both the total and peak contact forces, and is most comfortable for the user, with the assumption that the head remains stationary.

\section{Conclusion and Future Work}
This paper presents a novel approach for modeling physical human-robot interaction in simulation, addressing a gap in assistive robotics. By leveraging MuJoCo's accurate contact dynamics, we developed a soft-body human head model with an articulated mandible. Our integration of the flexible head model with a rigid skull captures the internal dynamics between the skin and the underlying skeleton, enabling realistic motion actuation and force transmission.

Using this head model that accounts for soft-body contacts, we could capture detailed contact interactions between a spoon and the mouth during bite transfer. This capability enables systematic evaluation of bite transfer parameters, offering valuable insights into how variations in bite transfer parameters like insertion depth, entry angle, exit depth, and exit angle influence contact forces. These insights are important in improving user safety and comfort in robot-assisted feeding. 

Nevertheless, the current approach assumes a static head during bite transfer, which overlooks the natural movements of the head in response to the spoon. Furthermore, the current model lacks a detailed oral cavity, tongue and lip actuation, which limits the realism of the simulated interactions. Future work will incorporate dynamic head and neck movements, and validate the simulation forces with real-world data.

\bibliographystyle{IEEEtran}
\bibliography{IEEEabrv, references}

\end{document}